\RequirePackage{fix-cm}
\documentclass[default,iicol]{sn-jnl}

\usepackage{fix-cm}
\usepackage{graphicx}
\usepackage{amsmath,amssymb}
\usepackage{multirow}
\usepackage{xcolor}
\usepackage{bm}
\usepackage{mathrsfs}
\usepackage{pifont}
\usepackage{ragged2e}
\usepackage{makecell}
\usepackage{colortbl}
\usepackage{rotating}
\usepackage{dcolumn}
\usepackage{threeparttable}
\usepackage{silence}
\usepackage{cleveref}
\usepackage{amsmath}
\usepackage{lmodern}

\usepackage{lmodern}
\usepackage{anyfontsize}
\usepackage[dvipsnames]{xcolor}

\newcommand{\para}[1]{\vspace{.05in}\noindent\textbf{#1}\quad}

\graphicspath{{./Imgs/}{./Imgs/Authors/}}
\DeclareGraphicsExtensions{.pdf,.jpg,.png}

\newcommand{\figref}[1]{Fig.~\ref{#1}}
\newcommand{\tabref}[1]{Tab.~\ref{#1}}
\newcommand{\secref}[1]{Sec.~\ref{#1}}

\newcommand{\equref}[1]{Eq.~(\ref{#1})}

\def\ie{\emph{i.e.}}

\jyear{2025}%

\theoremstyle{thmstyleone}%
\theoremstyle{thmstyletwo}%

\theoremstyle{thmstylethree}%

\begin{document}

\title[Article Title]{Evaluating SAM2 for Video Semantic Segmentation}

\author[1,2]{\fnm{Syed Ariff Syed} \sur{Hesham}}

\author[3]{\fnm{Yun} \sur{Liu}}
\equalcont{Corresponding authors.}

\author[3]{\fnm{Guolei} \sur{Sun}}
\equalcont{Corresponding authors.}

\author[4]{\fnm{Jing} \sur{Yang}}

\author[5]{\fnm{Henghui} \sur{Ding}}

\author[2]{\fnm{Xue} \sur{Geng}}

\author[1]{\fnm{Xudong} \sur{Jiang}}

\affil[1]{\orgdiv{School of EEE}, \orgname{Nanyang Technological University}, \orgaddress{\city{Singapore} \postcode{639798},  \country{Singapore}}}

\affil[2]{\orgdiv{Institute for Infocomm Research}, \orgname{A*STAR}, \orgaddress{\city{Singapore} \postcode{138632}, \country{Singapore}}}

\affil[3]{\orgdiv{College of Computer Science}, \orgname{Nankai University}, \orgaddress{\city{Tianjin} \postcode{300350},  \country{China}}}

\affil[4]{\orgdiv{State Key Laboratory of Public Big Data}, \orgname{Guizhou University}, \orgaddress{\city{Guiyang} \postcode{550025},  \country{China}}}

\affil[5]{\orgdiv{Fudan Vision and Learning Lab}, \orgname{Fudan University}, \orgaddress{\city{Shanghai} \postcode{200438},  \country{China}}}

\abstract{The Segmentation Anything Model 2 (SAM2) has proven to be a powerful foundation model for promptable visual object segmentation in both images and videos, capable of storing object-aware memories and transferring them temporally through memory blocks. While SAM2 excels in video object segmentation by providing dense segmentation masks based on prompts, extending it to dense Video Semantic Segmentation (VSS) poses challenges due to the need for spatial accuracy, temporal consistency, and the ability to track multiple objects with complex boundaries and varying scales. This paper explores the extension of SAM2 for VSS, focusing on two primary approaches and highlighting firsthand observations and common challenges faced during this process. The first approach involves using SAM2 to extract unique objects as masks from a given image, with a segmentation network employed in parallel to generate and refine initial predictions. The second approach utilizes the predicted masks to extract unique feature vectors, which are then fed into a simple network for classification. The resulting classifications and masks are subsequently combined to produce the final segmentation. Our experiments suggest that leveraging SAM2 enhances overall performance in VSS, primarily due to its precise predictions of object boundaries.
}


\maketitle

\section{Introduction}\label{sec:intro}
Segmentation is a fundamental task in computer vision that involves identifying and extracting the region of interest from a given image, representing an object of interest such as a person, barrier or road. A specialized task in this field is Video Semantic Segmentation (VSS) \cite{VSPW_dataset}, which focuses on classifying and segmenting each pixel in a video sequence into predefined categories over time. Unlike traditional image segmentation, which deals with static images, VSS accounts for the temporal dimension of video data, allowing for the identification and tracking of objects across multiple frames. This involves not only recognizing objects but also maintaining their identities and boundaries as they move, change shape, or interact with other elements in the scene \cite{CFFM,sun2024learning}.

Recently, there has been an increasing interest in applying foundation models in numerous visual tasks like classification, detection, and VSS. The excitement in the usage of foundation models originates from their extensive pre-training on large web-scale datasets, which in turn enhances their ability to generalize effectively to various downstream tasks. The commercial success of ChatGPT, which uses the foundation model GPT \cite{brown2020language} for text generation, has accelerated the exploration of foundation models in other modalities, including the field of computer vision \cite{zhai2022scaling,dehghani2023scaling,liu2022swin,wang2023videomae}. The early developments of the foundation model in computer vision include CLIP \cite{radford2021learning}, which makes use of a vast collection of image-text pairs from the web to align both modalities effectively. This was then applied to recognize new visual categories through text prompts and provided zero-shot generalization for vision concepts, ultimately inspiring the creation of the foundation model for object detection called GLEE \cite{wu2024general}. Meta AI's introduction of the Segment Anything Model (SAM) \cite{kirillov2023segment} for promptable segmentation, further broadened the application of foundation models across multiple domains, including entertainment, medicine, and forensics.

While SAM excels at segmenting static images, its effectiveness in real-world scenarios is restricted due to its inability to address the temporal information in streaming images/video data. To overcome this challenge, Meta AI developed SAM2 \cite{ravi2024sam}, an enhanced version of the original SAM that incorporates a memory module. This enhancement allows the storage of context memories for each generated masklet, allowing SAM2 to process video frames sequentially while referencing its prior memories of the objects of interest, which leads to a coherent segmentation of these objects over time. Together, SAM and SAM2 have sparked considerable interest among researchers, motivating investigations into SAM's potential for various applications and downstream tasks. Research has explored SAM's segmentation abilities across different contexts, including the medical domain and its role as a semantic segmentor \cite{yang2025unimatch, benigmim2024collaborating, xiong2024sam2, wu2023medical, he2024lightweight}. However, there has been limited focus on evaluating the capabilities of SAM2 in the context of VSS, particularly in challenging scenarios.

In this paper, we extend SAM2's application specifically to VSS in two main ways. First, we evaluate SAM2 as a post-processing refiner to extract different objects in a scene as masklets and then use them to enhance the segmentation quality of existing models, assessing the resulting improvements. Second, we explore SAM2's independent functionality as a semantic segmentor, where we utilize features from its image encoder to vectorize the representation of each detected object in an image and perform classification. Our experiments are conducted on the established VSPW dataset \cite{VSPW_dataset}, which includes a broad spectrum of short videos showcasing various motion levels, allowing for a thorough assessment of SAM2's capabilities in this specific task. 

Overall, the contributions of this paper include:
\begin{enumerate}
    \item We evaluate the effectiveness of SAM2 as a post-processing refiner to improve the quality of masks generated by VSS models across multiple datasets, identifying areas for potential enhancement.
    \item We investigate the standalone functionality of SAM2 to work independently (I-SAM2) as a video semantic segmentor, focusing on its performance with minimal modifications to the base model.
    \item We assess SAM2's performance using VSS benchmarks to identify its limitations and applicability in this domain, providing insights into the adaptation requirements of foundation models for dense prediction tasks.
\end{enumerate}

\section{Related Work}\label{sec:related_works}
Segmentation is a key challenge in computer vision, focusing on pixel-wise classification of images. It plays a crucial role in interpreting image information and is primarily explored through three main approaches: semantic segmentation, instance segmentation, and panoptic segmentation. Semantic segmentation \cite{chen2014semantic, hesham2021pix2pt, chen2017rethinking} involves assigning each pixel to a predefined category. Whereas, instance segmentation \cite{liu2018path,  bolya2019yolact} takes this a step further by separating segments into individual instances within the same class. Panoptic segmentation, as introduced in \cite{kirillov2019panoptic}, goes beyond both semantic and instance segmentation to provide a comprehensive understanding of the entire scene. While these techniques lay the foundation for image segmentation, the complexity of video data introduces additional challenges and opportunities for research, as VSS extends the principles of pixel-wise classification into the temporal domain to provide coherent segmentation across video frames.

\subsection{Video Semantic Segmentation}
The emergence of deep learning, particularly with the introduction of convolutional networks, has promoted extensive research focused on segmentation and its applications across various fields, including robotics \cite{ainetter2021end, li2019online, siam2019video, song2020grasping, pradeep2022self}, autonomous driving \cite{codevilla2018end,siam2018comparative, muhammad2022vision, xiao2023baseg}, and medical image analysis \cite{cciccek20163d, ronneberger2015u,grammatikopoulou2024spatio,wu2022semi,li2025semi,wang2021noisy}. While there has been a trend toward developing models specialized in segmentation, most approaches are developed to process images independently \cite{FCN, mobilenetv2, DeepLab, ronneberger2015u, PSPNet, yan2020roboseg}. This independent processing presents limitations when applying image-based models to the real world, which is inherently continuous. Consequently, there is a growing interest in developing models specifically designed for videos, allowing them to incorporate temporal information and ensure smoother and more consistent predictions over time \cite{qiu2017learning, CFFM, MRCFA, li2018low, paul2020efficient,hu2020temporally, shelhamer2016clockwork, zhu2019improving}. This increased focus on video segmentation has led to significant advancements that have positively impacted the overall performance of models in this domain \cite{zhou2022survey, gao2023deep, pan2024moda, guo2024vanishing, mai2024pay, sun2024learning,MPVSS, gao2023exploit, TV3S}, including recent modular architectures that leverage foundation models \cite{lee2025cavis, zhang2023dvis, zhang2025dvis++}. However, the demand for further performance improvements remains crucial. Therefore, it becomes essential to explore the segmentation using the latest techniques to continue enhancing performance.

\begin{figure*}[!t]
    \centering
    \includegraphics[width=\linewidth]{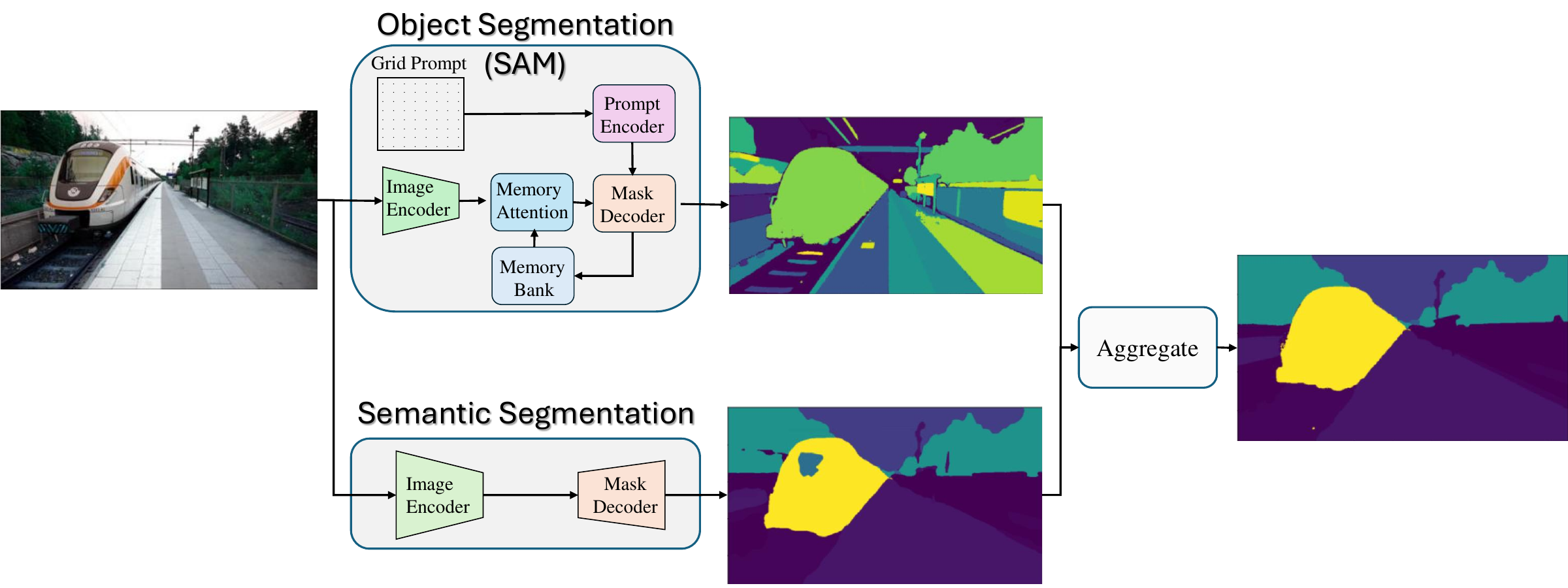}
    \caption{Framework for using SAM2 \cite{ravi2024sam} as a refiner to post-process the predicted masks from a segmentation model.}
    \label{fig:ourmodel2}
\end{figure*}

\subsection{Segment Anything} \label{ssec:Re_SAM}

Recently, there has been a growing interest in generalized AI frameworks, commonly termed \textit{foundational models} \cite{bommasani2021opportunities}. These models make use of web-scale datasets coupled with a self-supervised learning approach to achieve powerful zero-shot generalization across a wide range of downstream tasks. Significant advancements in Natural Language Processing (NLP) leading to models like BERT \cite{devlin2019bert} and GPT series \cite{brown2020language, achiam2023gpt, hurst2024gpt, jaech2024openai} have inspired the computer vision field to create Large Visual Models (LVMs) \cite{zhai2022scaling, dehghani2023scaling, liu2022swin}. Additionally, research integrating visual and textual data has produced models like CLIP \cite{radford2021learning} and ALIGN \cite{jia2021scaling}, making multimodal AI an exciting field. In line with the current trends, a significant contribution in this domain is Meta's Segment Anything Model (SAM) \cite{kirillov2023segment} which aims to provide a versatile, promptable solution for image segmentation tasks. 

While SAM excels in zero-shot generalization, it faces challenges when applied to video tasks due to the need for efficient frame-to-frame associations dealing with the increased dimensions, handling the temporal information and maintaining temporal consistency. Several methods, such as Track Anything Model (TAM) \cite{yang2023track} and SAM-Track \cite{cheng2023segment} for object tracking, and SAMText \cite{he2023scalable} for video text spotting, have attempted to address these challenges. However, they often require adjustments to the model architecture for optimal performance. SAM2 \cite{ravi2024sam} builds upon the original SAM \cite{kirillov2023segment} to tackle these limitations. It achieves this by integrating a streaming memory and further refines the segmentation by making use of interactive user prompts coupled with a memory attention mechanism, which allows the system to track target objects across frames, allowing for real-time video segmentation. As a result, SAM2 not only enhances the accuracy of segmentation but also improves overall efficiency with fewer user interactions, allowing for more effective segmentation in spatio-temporal contexts. The generalizability of SAM2 to various complex real-world scenarios is achieved through training on the SA-V dataset, a significantly larger and more diverse expansion of the dataset that was used to train SAM, comprising an impressive 50.9K videos and 652.6K masklets.

\subsubsection{Video Applications}

The introduction of SAM2 \cite{ravi2024sam} has led to a wave of research focused on various applications, including video generation \cite{lu2023can, wang2024disco, qin2023dancing}, 3D reconstruction \cite{yang2023sam3d, dong2023leveraging}, deepfake detection \cite{lai2023detect}, video editing \cite{wu2023cvpr, yin2023or}, and model compression \cite{zhao2023fast}. Its applications in Video Object Segmentation (VOS) covers areas like Video Object Tracking (VOT) \cite{chu2024zero, yang2023track}, Audio-Visual Segmentation \cite{mo2023av}, and domain-specific applications, including medical video segmentation \cite{yue2024surgicalsam}, among others \cite{benjdira2023rosgpt_vision, fu2024sam}. 

While SAM has been applied onto different use cases, there has been limited work done on a comprehensive evaluation of SAM2 in real-world video scenarios. This evaluation is crucial for assessing its performance and identifying the factors that influence its effectiveness as both a post-processing method and a standalone semantic segmentor. Therefore, this work aims to evaluate SAM2's performance in real-world scenarios, conducting thorough tests on the various parameters to report its direct practical applicability to images captured in the wild.

\section{Methods \& Experiments}\label{sec:methodology}
This section provides an overview of the experiments conducted, outlining the datasets used, the evaluation metrics employed, and the specific implementation settings applied throughout the study.

\subsection{Methods}

To evaluate the effectiveness of SAM2 \cite{ravi2024sam} as both a post-processing refiner and a standalone tool for generating segmentation masks, the automatic mask generator is used with a 16x16 point grid as input prompts. Furthermore, to evaluate the SAM2's ability to refine masks while ensuring new objects are tracked, a window of 32 frames is used to share and propagate masklets before they are recomputed. In the experiments, thresholds for the predicted intersection-over-union (IoU), stability, and stability offset are established at 0.6, 0.8, and 0.9, respectively, based on the ablations studies and visual assessments discussed in \secref{sssec:ab_refiner}. A detailed discussion of the evaluation setup for SAM2 in both contexts is covered in this section.

\begin{figure*}[!t]
    \centering
    \includegraphics[width=\linewidth]{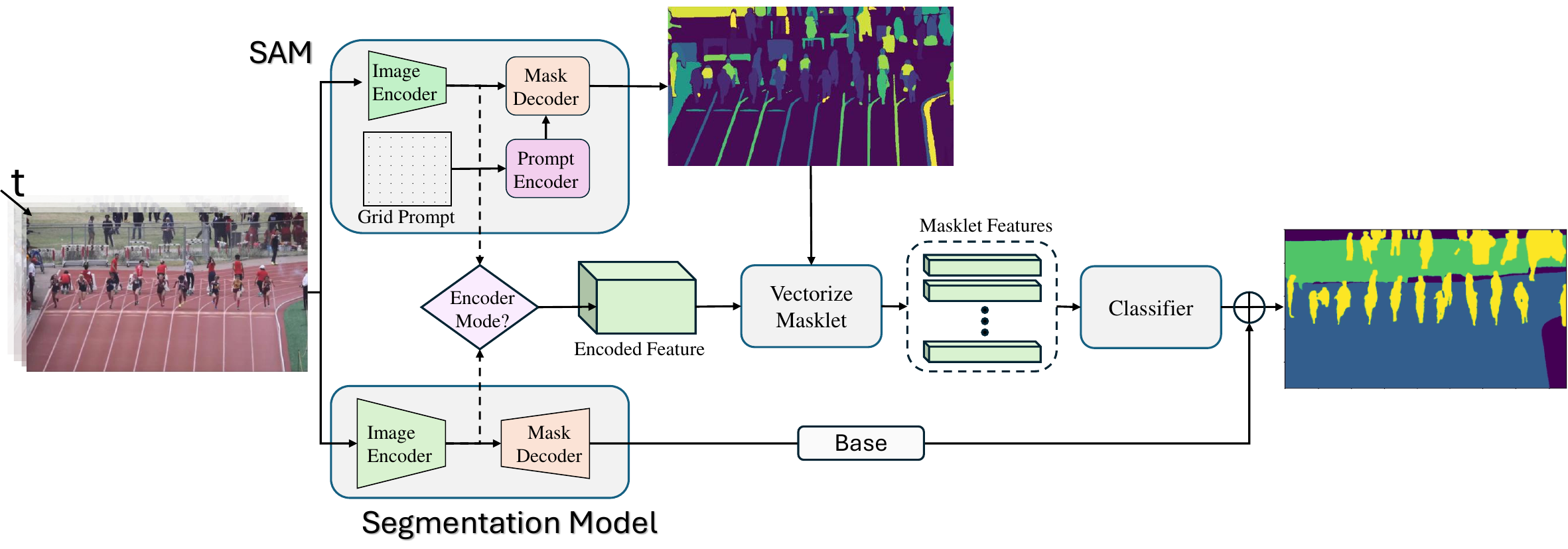}
    \caption{Framework for using SAM2 \cite{ravi2024sam} as an independent semantic segmentation model.}
    \label{fig:ourmodel1}
\end{figure*}

\subsubsection{SAM2 as refiner} \label{sssec:sam2refiner}
Using SAM2 as a refiner serves as a qualitative study to evaluate its effectiveness as a post-processing algorithm for refining masks produced by a trained segmentation model. The aim is to assess whether SAM2's capabilities as a generic foundation model can effectively extract unique objects from a scene, thereby potentially enhancing the uniformity and accuracy of the predicted masks through improved detection and refinement of object-level details.

The approach on using SAM2 as a refiner is illustrated in \figref{fig:ourmodel2}. In this method, a semantic segmentation model generates a semantic mask $S$, while SAM2 operates in parallel as an object segmentation network that extracts individual objects from the scene as masklet $M' = \{m_1, m_2, \dots, m_n\}$. After passing through an aggregator, each masklet $m_i$ is used to refine the predicted semantic mask by determining the predominant class $c_i$ within the masklet. The refined segmentation mask $R$ is then computed as follows:
\begin{equation} \label{eq:final_combiner}
{R(x, y) = \begin{cases}
c_i & \text{if } (x, y) \in m_i, \\
S(x, y) & \text{otherwise}.
\end{cases}    }
\end{equation}
Here, $S(x, y)$ is the original semantic mask at pixel location $(x, y)$ and $c_i$ represents the predominant class within the masklet $m_i$. To determine $c_i$, we compute the class label that appears most frequently within the mask $m_i$:
\begin{equation}
\label{eq:ci}
    {c_i = \arg\max_{c \in \mathcal{C}} \sum_{(x, y) \in m_i} \delta(S(x, y),c) }.
\end{equation}
Here, $\delta(S(x, y),c)$ is the Kronecker delta function, which returns 1 if $S(x,y)=c$ and 0 otherwise. Applying these equations across all the pixels produces the final refined mask \( R\). Evaluations are then performed on this refined mask $R$ to evaluate how the performance changes after using SAM2 as a post-processing unit.

\subsubsection{SAM2 as Segmentor} \label{sssec:sam2segmentor}
To investigate SAM2's potential as an independent semantic segmentor (I-SAM2) with minimal architectural modifications, a framework incorporating only the essential changes is employed, as illustrated in \figref{fig:ourmodel1}. The primary change involves the use of a simple classifier composed of a Multi-Layer Perceptron (MLP) layer. The process of obtaining a segmentation mask begins with the images or live-stream frames as input, which are encoded by SAM2's image encoder. The encoded features $F$ with feature dimension $d$ are subsequently processed through a mask decoder, resulting in the generation of masklets \( M = \{m_1, m_2, \dots, m_n\} \) that represent $n$ detected object regions in the scene. A mean of the encoded features corresponding to the area covered by each masklet is computed to derive a representation for each masklet. This yields a single vector that characterizes each mask which can be expressed as:
\begin{equation} \label{eq:vectorize}
{v_i = \frac{1}{\vert m_i\vert}\sum_{(x,y)\in m_i} F(x,y)}.
\end{equation}
Here, $v_i\in\mathbb{R}^d $ represents a feature vector of $i$-th mask ($i\in\{1,2 ... n\}$), $F(x,y)\in\mathbb{R}^d$ is the encoded feature vector at pixel location $(x,y)$, and $\vert m_i\vert$ represents the cardinality, \ie, the number of pixels covered by the mask $m_i$. 

The feature vectors \( \{v_1, v_2, \dots, v_n\} \) are subsequently fed into the classifier for class prediction of each vector.
The outputs from this classification process are combined with the masklets to create a unified segmentation mask, which is then used to assess the performance of SAM2 as a standalone semantic segmentor.

To further investigate the capabilities of SAM2 and compare its effectiveness with semantic information encoding, evaluations are conducted on other VSS methods by using their encoders to extract encoded features for training independent classifiers. Following the methodology outlined in \equref{eq:vectorize}, the encoded features from the trained segmentation models are extracted and used to train the classifiers for evaluation purposes.

Additionally, it was observed that the detection of masklets within the images occasionally resulted in low mask coverage, leaving significant areas without predicted labels. To address this issue, a base model is introduced, which fills areas not covered by the masklets with predictions from the base segmentor, as shown in  \figref{fig:ourmodel1}. 
The incorporation of a base model aids in providing a thorough evaluation across diverse scenarios. Further insights and discussions regarding the evaluations and their results can be found in \secref{ssec:res_segmentor}.

\subsection{Dataset}\label{ssec:dataset}
There are numerous datasets available for Video Object Segmentation (VOS) and Video Instance Segmentation (VIS), including one from SAM2 \cite{xiong2024sam2}. However, there is a limited availability of densely annotated datasets for evaluating semantic segmentation, with the VSPW dataset \cite{VSPW_dataset} being a notable exception. The VSPW dataset stands out by providing dense annotations at a frame rate of 15 fps and includes 124 unique categories. It contains 2,806 training clips (which totals 198,244 frames), 343 validation clips (24,502 frames), and 387 test clips (28,887 frames), and it has been recognized by the community as a benchmarking dataset for VSS. We chose to use the VSPW dataset for our experiments as it covers a wide range of indoor and outdoor scenes, making it particularly suitable for performance evaluation.

\subsection{Performance Metrics}\label{ssec:metrics}
A well established metric for semantic segmentation, mean IoU (mIoU) is used for qualitative assessment of the experiments. Considering the characteristics of SAM2 as a generic foundation model suitable for video object segmentation, the evaluation extends beyond mIoU to include additional metrics to assess visual consistency across frames and the precision of detected object boundaries. Such multidimensional evaluation provides with deeper insights into the model's effectiveness, ensuring that the segmentation results maintain high fidelity over temporal video sequences. Firstly, to assess the visual consistency of the predictions, the metric mean Video Consistency (mVC) \cite{VSPW_dataset} is used. The metric mVC specifically measures the temporal smoothness of the predicted segmentation maps across frames, assessing performance in the temporal domain. Formally, for a video clip represented as \(\{\bm{I}_{c}\}^{C_v}_{c=0}\), with corresponding ground-truth masks \(\{\bm{M}_{c}\}^{C_v}_{c=1}\) and predicted outputs \(\{\bm{O}_{c}\}^{C_v}_{c=1}\), the calculation of video consistency VC\(_n\) is performed as follows:

\footnotesize
\begin{equation} \label{eq:VC}
\text{VC}_n = \frac{1}{{C_v}-n+1} \sum_{i=1}^{{C_v}-n+1} \frac{\left( \bigcap_{i}^{i+n-1} \bm{M}_i \right) \cap \left( \bigcap_{i}^{i+n-1} \bm{O}_i \right)}{\bigcap_{i}^{i+n-1} \bm{M}_i}.
\end{equation}
\normalsize
The final mVC\textsubscript{n} is computed by taking the mean of all the calculated VC\(_n\) values, given that \( C\textsubscript{v} \geq n \). As shown in \equref{eq:VC}, mVC\textsubscript{n} captures the common areas of the predicted masks across frames, reflecting the consistency of these predictions over time. For further details on this metric, please refer to \cite{VSPW_dataset}.

Secondly, to evaluate the accuracy of object boundaries, the ground-truth masks \( M \), consisting of $C$ classes,  are processed using morphological operations with a 5-pixel kernel $K$. This processing results in the extraction of boundaries for each class $c \in C$, resulting in a boundary mask $B$. Subsequently, the IoU is computed between these boundary masks $B$ and the predicted output masks $O$. An average of the IoU values across $N$ images in the dataset provides with the mean boundary IoU, referred to as \(\text{mBIoU}\). This process is formulated as:
\begin{equation}
B = \bigcup_{c \in C} \left( M_c \ominus K \right),
\end{equation}
\footnotesize
\begin{equation} \label{eq:miou_border}
{\text{mBIoU} = \frac{1}{N} \sum_{i=1}^{N} \frac{\vert (M_i \cap O_i) \cap B \vert}{\vert M_i \cap B \vert + \vert O_i \cap B \vert - \vert (M_i \cap O_i) \cap B \vert}}.
\end{equation}
\normalsize
The above \equref{eq:miou_border} acts as a comprehensive measurement for the accuracy of borders for the predicted segmentation masks, which helps in evaluating the alignment of the predicted boundaries with the ground truth. The metric will be particularly beneficial in applications, where the precise delineation of object edges is of critical importance, such as in medical imaging, autonomous driving, and various other computer vision tasks.

\begin{table*}[!t]
  \centering
  \setlength{\tabcolsep}{4.8mm}
  \caption{Quantitative comparison of our model with existing methods on the VSPW dataset \cite{VSPW_dataset} with an input resolution of 480 × 853.}
  \label{tab:VSPW_comparision}
  \resizebox{\linewidth}{!}{
  \begin{tabular}{l|c|c|c|c|c|c}\toprule
    \textbf{Methods} & \textbf{Backbones} &\textbf{ mIoU$\uparrow$} & \textbf{Weighted IoU$\uparrow$} & \textbf{mBIoU$\uparrow$} & \textbf{mVC\textsubscript{8}$\uparrow$} & \textbf{mVC\textsubscript{16}$\uparrow$} \\
    \midrule
    Segformer \cite{xie2021Segformer}   & MiT-B0        & 32.9           & 56.8  & -& 82.7 & 77.3 \\
    CFFM \cite{CFFM}                    & MiT-B0        & 35.4           & 58.5  & 21.8
& 87.7 & 82.9 \\
    MRCFA \cite{MRCFA}                  & MiT-B0        & 35.2           & 57.9  & 21.5& 88.0 & 83.2 \\
    SAM2-CFFM& MiT-B0        & 36.7& \textbf{60.0}& \textbf{25.8}& 88.8& 84.4\\
    SAM2-MRCFA& MiT-B0        & \textbf{36.8}& 59.2& \textbf{25.8}& \textbf{89.1}& \textbf{84.6}\\
    \midrule
    
    Segformer \cite{xie2021Segformer}   & MiT-B1        & 36.5           & 58.8  & -& 84.7 & 79.9 \\
    CFFM \cite{CFFM}                    & MiT-B1        & 38.5           & 60.0  & 24.7& 88.6 & 84.1 \\
    MRCFA \cite{MRCFA}                  & MiT-B1        & 38.9           & 60.0  & 24.7& 88.8 & 84.4 \\
    TV3S \cite{TV3S} & MiT-B1        & 40.0  & -     & 25.6& 90.7 & 87.0 \\ 
    SAM2-CFFM& MiT-B1        & 40.3& 61.1& 29.3& 89.4& 85.1\\
    SAM2-MRCFA& MiT-B1        & 40.6& 60.9& 29.4& 89.5& 85.2\\
    SAM2-TV3S& MiT-B1        & \textbf{41.1}& \textbf{62.3}& \textbf{30.1}& \textbf{91.4}& \textbf{87.8}\\
    \midrule
    
    Segformer \cite{xie2021Segformer}   & MiT-B2        & 43.9           & 63.7  & -& 86.0 & 81.2 \\
    CFFM \cite{CFFM}                    & MiT-B2        & 44.9           & 64.9  & 28.9& 89.8 & 85.8 \\
    MRCFA \cite{MRCFA}                  & MiT-B2        & 45.3           & 64.7  & 28.7& 90.3 & 86.2 \\
    TV3S \cite{TV3S} & MiT-B2        & 46.3  & -     & 29.8& 91.5 & 88.4\\ 
    SAM2-CFFM& MiT-B2        & 46.7& 65.3& 33.9& 90.4& 86.5\\
    SAM2-MRCFA& MiT-B2        & 46.9& \textbf{65.7}& 33.8& 91.0& 87.1\\
    SAM2-TV3S& MiT-B2        & \textbf{47.8}& 65.2& \textbf{34.8}& \textbf{92.1}& \textbf{89.1}\\
    \midrule
    Segformer \cite{xie2021Segformer} & MiT-B5 & 48.9  & 65.1  & -& 87.8 & 83.7 \\
    CFFM \cite{CFFM} & MiT-B5 & 49.3  & 65.8  & 32.1& 90.8 & 87.1 \\
    MRCFA \cite{MRCFA} & MiT-B5 & 49.9  & 66.0  & 32.0& 90.9 & 87.4 \\
    TV3S \cite{TV3S} & MiT-B5 & 49.8  & -  & 32.9& 91.7 & 88.7 \\ 
    SAM2-CFFM& MiT-B5        & 50.8& 67.3& 37.0& 91.2& 87.7\\
    SAM2-MRCFA& MiT-B5        & \textbf{51.3}& 66.6& 37.1& 91.5& 88.2\\
    SAM2-TV3S & MiT-B5        & \textbf{51.3}& \textbf{67.1}& \textbf{38.1}& \textbf{92.2}& \textbf{89.3}\\
    \bottomrule
  \end{tabular}}
\end{table*}

\section{Results and Ablations}\label{sec:observations}

This section provides a comprehensive analysis of the results obtained from various experiments conducted on SAM2, aimed at applying it to the semantic segmentation task. All experiments were performed using the VSPW dataset \cite{VSPW_dataset}. The following discussion will present the findings and offer insights into the interpretation of the results obtained.

\subsection{Results as Refiner}

\begin{figure*}[!t]
    \centering
    \includegraphics[width=\linewidth]{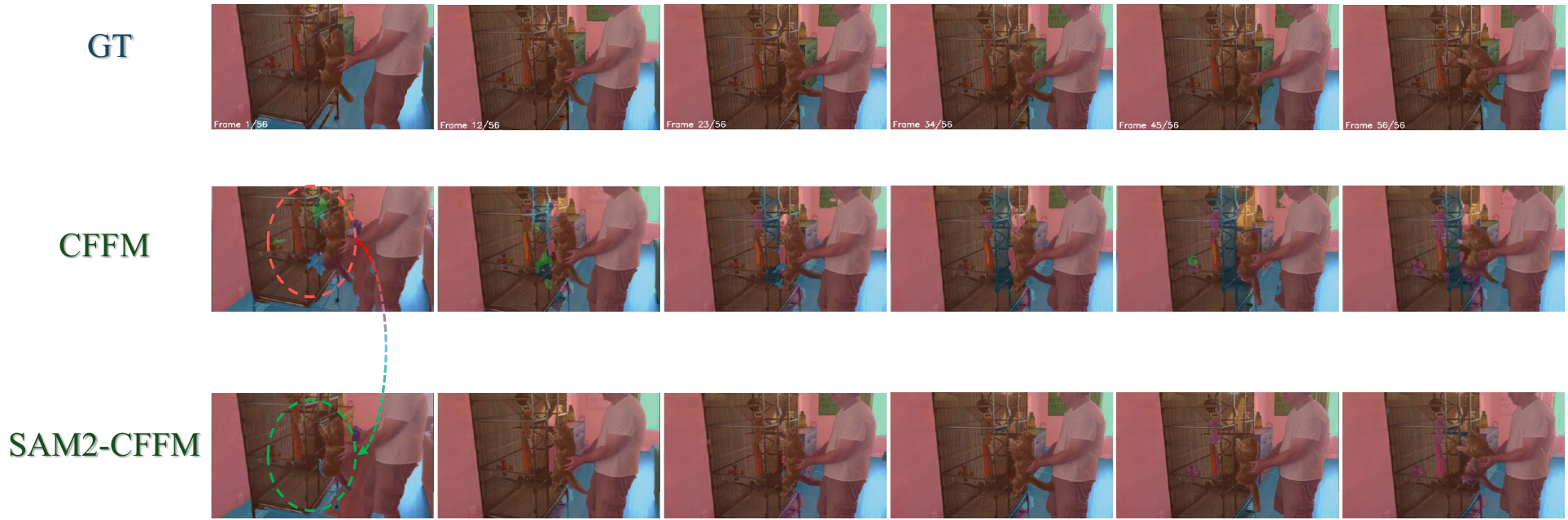}
    \caption{Qualitative comparison of SAM2 refinement on VSPW validation set. SAM2 refinement improves boundary precision and reduces noise in predicted masks, particularly for objects with complex boundaries.}
    \label{fig:qualitative}
\end{figure*}

The performance of SAM2 as a mask refiner is evaluated using state-of-the-art VSS models, following the framework illustrated in \figref{fig:ourmodel2}. The evaluation results presented in \tabref{tab:VSPW_comparision} indicate that SAM2 is effective as a refiner, enhancing the overall performance of the models across all evaluation metrics. Furthermore, SAM2 shows significant improvements in object segmentation as reflected by the notable increase in mBIoU scores. A consistent improvement in visual consistency across various VSS backbones and their variants (MiT-B0 to MiT-B5) can be attributed to SAM2's ability to treat objects as masklets and effectively track them across frames. The consistent enhancements across different evaluation metrics underscore SAM2's effectiveness as a post-processing tool, particularly in refining object boundaries and improving overall segmentation accuracy and consistency. Qualitative results in \figref{fig:qualitative} illustrate these improvements, showing that SAM2 refines object boundaries and reduces segmentation noise across diverse scene categories.

\para{Computational Cost Analysis.}
We evaluate the computational overhead of the SAM2-refiner on the VSPW validation set using a single NVIDIA A100 GPU. The refiner operates at 1.1 to 3.5 FPS depending on scene complexity and the number of objects tracked, compared to 4 FPS to $\sim$15 FPS for base to large VSS models. This computational cost positions SAM2 as most suitable for offline refinement scenarios where segmentation quality, particularly boundary accuracy and temporal consistency, is prioritized over real-time performance.

\subsubsection{Cross-Dataset Generalization}

To evaluate the generalization capability of SAM2 as a refiner across different datasets and scenes, we conduct experiments on VIPSeg~\cite{miao2022large}, a large-scale video panoptic segmentation dataset with 124 classes covering diverse in-the-wild scenarios. We apply the SAM2-refiner to predictions from recent state-of-the-art video panoptic segmentation methods by converting their panoptic outputs to semantic segmentation masks. Specifically, we evaluate CAVIS \cite{lee2025cavis}, DVIS \cite{zhang2023dvis}, and DVIS++ \cite{zhang2025dvis++} with different backbones without any fine-tuning or parameter adjustment to the SAM2-refiner. The results in \tabref{tab:cross_dataset} demonstrate that SAM2 consistently improves boundary accuracy as indicated by mBIoU gains across all methods, while maintaining competitive performance on other metrics. This validates that the refinement capability of SAM2 generalizes well to diverse datasets and modern architectures without dataset-specific tuning.

\subsubsection{Ablations on Refiner} \label{sssec:ab_refiner}

Ablation studies are conducted to analyze the impact of various parameters during evaluation and their effects on the final segmentation results. All ablation experiments are conducted by making use of the MiT-B1 backbone \cite{xie2021Segformer} of the CFFM \cite{CFFM} VSS model, following the default parameters outlined in \secref{sec:methodology}. The ablations are conducted on four parameters that significantly influence model performance during inference: the temporal window size for mask propagation, the number of point grid inputs used as prompts, and the thresholds for prediction confidence and masklet stability. Understanding and testing these parameters is crucial for optimizing the model's effectiveness and ensuring reliable results.

\para{Effect of temporal window size.}
The evaluation of SAM2's ability to propagate masklets across sequential frames is performed by varying the temporal window size for masklet propagation. Using a temporal window allows for the detection of new objects while refreshing information about existing masklets, leading to more accurate segmentation. The results of the experiments are summarized in \tabref{tab:ab_frame}, where the evaluations begin with a single-frame temporal window and are incrementally increased up to a total of 32 frames. It is observed that, while increasing the number of frames has a negligible impact on the quantitative quality of the segmentation masks (mIoU), there is a consistent improvement in visual consistency. This improvement can be attributed to the tracking of masklets across the temporal dimension, which leads to more stable predictions of individual objects within the scene. Considering this observation, along with the advantage of faster inference times in SAM2 with increased temporal window, a temporal window of 32 frames is used for all subsequent experiments.

\begin{table}[!t]
\centering
\caption{Cross-dataset generalization of SAM2 as refiner on VIPSeg dataset.}
\label{tab:cross_dataset}
\resizebox{\columnwidth}{!}{%
\begin{tabular}{l|c|c|c|c}
\toprule
\textbf{Method} & \textbf{mIoU$\uparrow$} & \textbf{mBIoU$\uparrow$} & \textbf{mVC\textsubscript{8}$\uparrow$} & \textbf{mVC\textsubscript{16}$\uparrow$} \\
\midrule
CAVIS (R50)\cite{lee2025cavis} & 50.4 & 55.3 & 86.2 & 79.4 \\
DVIS++ (R50)\cite{zhang2025dvis++} & 49.9 & 55.5 & 85.8 & 79.5 \\
DVIS (Swin-L)\cite{zhang2023dvis} & 55.2 & 57.2 & 89.5 & 84.2 \\
DVIS (Swin-L)$^{\dagger} ~\cite{zhang2023dvis}$ & 57.0 & 60.9 & 91.4 & 87.7 \\
SAM2-CAVIS & 50.3 & 56.8 & 86.7 & 80.1 \\
SAM2-DVIS++ & 50.0 & 57.0 & 86.3 & 80.1 \\
SAM2-DVIS & 55.0 & 59.1 & 89.8 & 84.6 \\
SAM2-DVIS(offline) & \textbf{57.2} & \textbf{62.3} & \textbf{91.8} & \textbf{88.3} \\
\bottomrule
\end{tabular}%
}
\end{table}

\begin{table*}[!t]
  \centering
  \setlength{\tabcolsep}{4.8mm}
  \caption{Quantitative comparison of our model with existing methods on the VSPW dataset \cite{VSPW_dataset} with an input resolution of 480 × 853.} 
  \label{tab:MLP_comparision}
  \resizebox{\linewidth}{!}{
  \begin{tabular}{l|c|c|c|c|c|c}\toprule
    \textbf{Methods} & \textbf{Backbones} & \textbf{mIoU$\uparrow$} & \textbf{Weighted IoU$\uparrow$}  &\textbf{mBIoU$\uparrow$} & \textbf{mVC\textsubscript{8}$\uparrow$} & \textbf{mVC\textsubscript{16}$\uparrow$} \\
    \midrule
    CFFM \cite{CFFM}                    & MiT-B0        & 35.4           & 58.5   &21.5& 87.7 & 82.9 \\
    I-CFFM& MiT-B0        & 29.9& 49.4 &21.1& 66.9& 58.1\\
    I-CFFM\textsubscript{base}& MiT-B0        & 33.5& 56.9 &21.4& 73.9& 66.8\\
    \midrule
    CFFM                                & MiT-B1        & 38.5           & 60.0   &24.7& 88.6 & 84.1 \\
    I-CFFM& MiT-B1        & 30.3& 49.9 &21.3& 66.0& 57.3\\
    I-CFFM\textsubscript{base}& MiT-B1        & 34.1& 57.6 &22.0& 73.2& 66.2\\
    \midrule
    CFFM                                & MiT-B2        & 44.9           & 64.9   &28.9
& 89.8 & 85.8 \\
    I-CFFM& MiT-B2        & 37.5& 55.1 &27.1& 70.8& 62.7\\
    I-CFFM\textsubscript{base}& MiT-B2        & 42.3& 63.2 &28.1& 79.1& 73.0\\
    \midrule
    I-SAM2& -& 18.03& 36.5 &13.3& 52.8& 43.7\\
    I-SAM2\textsubscript{base}& -& 20.9& 41.8&15.7& 55.5& 46.6\\
    \bottomrule
  \end{tabular}}
\end{table*}

\para{Influence of grid prompt size.}
To effectively segment all objects in a scene using an automatic prompt segmentor, selecting the appropriate size for the point grid to perform prompting is essential. Based on the evaluations in \tabref{tab:ab_ptgrid}, there are minimal differences in performance across various prompt sizes. While the $8\times 8$ option might seem appealing due to potentially faster inference times, this study focuses on identifying all objects in the scene with minimal omissions, ensuring an accurate assessment of semantic segmentation. Upon further visual inspection, the $16\times 16$ grid appeared to be the optimal choice, as it does not significantly impact evaluation time. However, in scenarios where rapid inference is critical and some object omission is acceptable, smaller prompts may provide advantages based on the observed results.

\para{Impact on the choice of thresholds.}
The impact of the confidence threshold has been thoroughly evaluated and documented in \tabref{tab:ab_confidence}, while assessments of stability are provided in \tabref{tab:ab_stability}. The selection of optimal thresholds for confidence and stability should be guided by the specific requirements of the downstream tasks that utilize the segmentation capabilities of SAM2. A high confidence threshold results in a limited number of masklets that align well with the objects, often leading to the exclusion of many masklets due to stringent thresholding, while a low confidence threshold can produce an overwhelming number of masklets, even if they are poorly aligned. In terms of stability, a high threshold ensures consistency among the masklets, whereas a low threshold may permit unstable masklets to be included. For the purpose of this study, aimed at achieving effective semantic segmentation, it is essential to obtain a sufficient number of masklets while maintaining the stability of the predictions. Thus, the threshold values are implemented with 0.6 for confidence and 0.8 for stability to establish an optimal balance.

\subsection{Results as Segmentor} \label{ssec:res_segmentor}

Using SAM2 for semantic segmentation shifts the focus from classifying each pixel into specified categories to detecting every individual item in the scene—both ``thing'' and ``stuff''—before classifying these detected objects with a classifier. As discussed in \secref{sssec:sam2segmentor}, an evaluation is conducted to compare the independent SAM2 (I-SAM2) with other VSS model, CFFM \cite{CFFM}, in a similar setting (I-CFFM). Additionally, to verify the implication of the missed masks, a baseline mode is employed with optimal thresholding for the current task, as discussed in \secref{sssec:ab_refiner}, where the regions not covered by the masklets are filled with the segmentation predictions. For the I-SAM2 base mode (I-SAM2\textsubscript{base}) the segmentation results from CFFM variant with the MiT-B1 backbone \cite{xie2021Segformer} is used. 

The evaluation, detailed in \tabref{tab:MLP_comparision} indicates that SAM2 performs poorly in this task, even with assistance from the segmentors in the base mode, which has the worst performance compared to other methods. This trend holds true even for the trained VSS models, as there is a noticeable decline in performance, with a drop of at least 10\% in the quantitative accuracy of predictions, measured by mIoU values. Furthermore, there is a significant decrease in the stability of the results, evidenced by a drastic decline in visual consistency. The drastic drop in performance in SAM2 strongly suggests that it is not suitable for independent semantic segmentation when the core SAM2 model is not retrained for the specific task.

\begin{table}[!t] 
\centering
\caption{Ablation study on varying temporal window sizes (\ie, the number of frames).}\label{tab:ab_frame}
\begin{tabular}{c|c|c|c|c}
\toprule
\textbf{\#Frames} & \textbf{mIoU} & \textbf{FWIoU} & \textbf{mVC\textsubscript{8}} & \textbf{mVC\textsubscript{16}} \\
\midrule
1 & 40.3& 61.1& 88.2& 83.9\\
2 & 40.3& 61.1& 88.4& 84.0\\
4 & 40.3& 61.1& 88.6& 84.3\\
8 & 40.3& 61.0& 88.9& 84.5\\
10 & 40.3& 61.0& 89.1& 84.7\\
16 & 40.4& 61.1& 89.2& 84.8\\
\textbf{32} & 40.4& 61.1& 89.4& 85.1\\
\bottomrule
\end{tabular}
\end{table}

\begin{table}[!t] 
\centering
\caption{Ablation study on the grid prompt size.}
\label{tab:ab_ptgrid}
\begin{tabular}{c|c|c|c|c}
\toprule
\textbf{Prompt} & \textbf{mIoU} & \textbf{FWIoU} & \textbf{mVC\textsubscript{8}} & \textbf{mVC\textsubscript{16}} \\
\midrule
8 x 8 & 39.9& 61.0& 89.3& 85.1\\
\textbf{16 x 16} & 40.4& 61.1& 89.4& 85.1\\
32 x 32 & 40.3& 61.1& 89.3& 85.0\\
64 x 64 & 40.2& 61.0& 89.2& 84.9\\
\bottomrule
\end{tabular}
\end{table}

\para{Analysis of I-SAM2 Limitations.}
The performance gap observed in I-SAM2 can be understood by examining the fundamental differences between object segmentation and semantic segmentation objectives. SAM2 was trained for promptable object segmentation, where the primary goal is to delineate individual objects regardless of their semantic category. Consequently, the features learned by SAM2's image encoder are optimized for capturing general object properties such as boundaries, textures, and spatial coherence rather than class-discriminative information. This contrasts with models like CLIP \cite{radford2021learning}, which explicitly align visual and textual representations to encode semantic meaning, or task-specific VSS encoders that are supervised with dense categorical labels. The limitations are further compounded by the feature aggregation strategy in \equref{eq:vectorize}, which computes a mean feature vector over each masklet assuming semantic homogeneity within detected regions. While this assumption holds for class-aware encoders, it becomes problematic for SAM2's class-agnostic representations where averaging may dilute any latent discriminative signals. These observations suggest that leveraging SAM2 for semantic segmentation benefits more from utilizing its precise boundary predictions as a refinement tool rather than relying on its feature representations for direct classification.

\begin{table}[!t]
\centering
\caption{Ablation study on the confidence threshold.}
\label{tab:ab_confidence}
\begin{tabular}{c|c|c|c|c}
\toprule
\textbf{IoU Conf.} & \textbf{mIoU} & \textbf{FWIoU} & \textbf{mVC\textsubscript{8}} & \textbf{mVC\textsubscript{16}} \\
\midrule
 0.4 & 40.1& 61.1& 89.3& 84.9\\
 0.5 & 40.2& 61.1& 89.3& 85.0\\
 \textbf{0.6} & 40.4& 61.1& 89.4& 85.1\\
 0.7 & 40.3& 61.1& 89.4& 85.1\\
 0.8 & 40.2& 61.0& 89.5& 85.2\\
 0.9 & 39.8& 60.8& 89.3& 85.1\\
\bottomrule
\end{tabular}
\end{table}

\begin{table}[!t] 
\centering
\caption{Ablation study on the stability.}
\label{tab:ab_stability}
\begin{tabular}{c|c|c|c|c}
\toprule
\textbf{Stability} & \textbf{mIoU} & \textbf{FWIoU} & \textbf{mVC\textsubscript{8}} & \textbf{mVC\textsubscript{16}} \\
\midrule
 0.4 & 40.5& 61.2& 89.3& 85.0\\
 0.5 & 40.5& 61.2& 89.3& 85.0\\
 0.6 & 40.4& 61.2& 89.3& 85.0\\
 0.7 & 40.4& 61.1& 89.3& 85.0\\
 \textbf{0.8} & 40.4& 61.1& 89.4& 85.1\\
 0.9 & 40.0& 60.9& 89.3& 85.0\\
\bottomrule
\end{tabular}
\end{table}

\section{Challenges and Future Directions}\label{sec:challenges}

The limitations observed in our standalone I-SAM2 experiments reveal fundamental challenges in adapting foundation models for dense VSS. As discussed in \secref{ssec:res_segmentor}, SAM2's class-agnostic features optimized for objectness struggle with semantic discrimination required for pixel-level classification. This finding highlights the need for targeted adaptations that inject semantic awareness while preserving SAM2's strong boundary prediction capabilities. To address this, specialized modules or adapters can enhance SAM2's ability to understand and track multiple objects with complex boundaries and varying scales. For instance, SAM2-Adapter \cite{chen2024sam2} has been developed to adapt SAM2 for downstream tasks, achieving state-of-the-art results in areas like camouflaged object detection and shadow detection.

Building on these insights, future work could explore lightweight parameter-efficient strategies to bridge the semantic gap we identified. Low-Rank Adaptation (LoRA) could selectively update attention weights to enable semantic discrimination without full fine-tuning of SAM2's encoder. Alternatively, learnable fusion modules could dynamically weight SAM2's object-aware features against task-specific semantic embeddings, replacing our fixed majority voting approach with adaptive integration that learns optimal combination strategies during training. Such approaches would directly address the feature-level limitations revealed in our experiments while maintaining SAM2's zero-shot generalization for novel object categories.

Beyond semantic adaptation, evaluating SAM2 in open-vocabulary segmentation tasks highlights additional opportunities for enhancement. While SAM2 shows strong zero-shot generalization, its performance in open-vocabulary segmentation can be further improved by incorporating composable prompts. The SAM-CP \cite{chen2025samcp} approach pairs SAM2 with composable prompts and has achieved superior results in panoptic segmentation. This suggests that augmenting SAM2 with language-driven prompting could enhance its ability to segment a wider range of object categories in diverse and dynamic environments, complementing the semantic adaptations discussed above.

\section{Conclusion}\label{sec:conclusion}
This study investigates the application of the Segmentation Anything Model 2 (SAM2) in VSS by evaluating it across multiple datasets and architectures. Our experiments on VSPW and VIPSeg datasets demonstrate that SAM2, as a mask refiner, significantly enhances semantic segmentation performance by improving the precision of object boundaries and enhancing visual consistency across various existing segmentation models. The consistent improvements in mBIoU across different methods and datasets validate SAM2's effectiveness as a post-processing refinement tool. However, when evaluated as a standalone segmentor, SAM2 performs subpar compared to models explicitly trained for semantic segmentation. This limitation stems from SAM2's class-agnostic feature representations optimized for objectness rather than semantic discrimination. These findings underscore the critical need to adapt foundation models to specific tasks to fully leverage their advanced capabilities, while highlighting SAM2's practical value as a quality-focused refinement module for VSS applications.

\bibliographystyle{IEEEtran}
\bibliography{main}

\begin{thebibliography}{10}
\providecommand{\url}[1]{#1}
\csname url@samestyle\endcsname
\providecommand{\newblock}{\relax}
\providecommand{\bibinfo}[2]{#2}
\providecommand{\BIBentrySTDinterwordspacing}{\spaceskip=0pt\relax}
\providecommand{\BIBentryALTinterwordstretchfactor}{4}
\providecommand{\BIBentryALTinterwordspacing}{\spaceskip=\fontdimen2\font plus
\BIBentryALTinterwordstretchfactor\fontdimen3\font minus \fontdimen4\font\relax}
\providecommand{\BIBforeignlanguage}[2]{{%
\expandafter\ifx\csname l@#1\endcsname\relax
\typeout{** WARNING: IEEEtran.bst: No hyphenation pattern has been}%
\typeout{** loaded for the language `#1'. Using the pattern for}%
\typeout{** the default language instead.}%
\else
\language=\csname l@#1\endcsname
\fi
#2}}
\providecommand{\BIBdecl}{\relax}
\BIBdecl

\bibitem{VSPW_dataset}
J.~Miao, Y.~Wei, Y.~Wu, C.~Liang, G.~Li, and Y.~Yang, ``{VSPW}: A large-scale dataset for video scene parsing in the wild,'' in \emph{{IEEE Conf. Comput. Vis. Pattern Recog.}}, 2021, pp. 4133--4143.

\bibitem{CFFM}
G.~Sun, Y.~Liu, H.~Ding, T.~Probst, and L.~Van~Gool, ``Coarse-to-fine feature mining for video semantic segmentation,'' in \emph{{IEEE Conf. Comput. Vis. Pattern Recog.}}, 2022, pp. 3126--3137.

\bibitem{sun2024learning}
G.~Sun, Y.~Liu, H.~Ding, M.~Wu, and L.~Van~Gool, ``Learning local and global temporal contexts for video semantic segmentation,'' \emph{{IEEE Trans. Pattern Anal. Mach. Intell.}}, vol.~46, no.~10, pp. 6919--6934, 2024.

\bibitem{brown2020language}
T.~Brown, B.~Mann, N.~Ryder, M.~Subbiah, J.~D. Kaplan, P.~Dhariwal, A.~Neelakantan, P.~Shyam, G.~Sastry, A.~Askell \emph{et~al.}, ``Language models are few-shot learners,'' \emph{Advances in neural information processing systems}, vol.~33, pp. 1877--1901, 2020.

\bibitem{zhai2022scaling}
X.~Zhai, A.~Kolesnikov, N.~Houlsby, and L.~Beyer, ``Scaling vision transformers,'' in \emph{{IEEE Conf. Comput. Vis. Pattern Recog.}}, 2022, pp. 12\,104--12\,113.

\bibitem{dehghani2023scaling}
M.~Dehghani, J.~Djolonga, B.~Mustafa, P.~Padlewski, J.~Heek, J.~Gilmer, A.~P. Steiner, M.~Caron, R.~Geirhos, I.~Alabdulmohsin \emph{et~al.}, ``Scaling vision transformers to 22 billion parameters,'' in \emph{{Int. Conf. Mach. Learn.}}, 2023, pp. 7480--7512.

\bibitem{liu2022swin}
Z.~Liu, H.~Hu, Y.~Lin, Z.~Yao, Z.~Xie, Y.~Wei, J.~Ning, Y.~Cao, Z.~Zhang, L.~Dong \emph{et~al.}, ``Swin transformer v2: Scaling up capacity and resolution,'' in \emph{{IEEE Conf. Comput. Vis. Pattern Recog.}}, 2022, pp. 12\,009--12\,019.

\bibitem{wang2023videomae}
L.~Wang, B.~Huang, Z.~Zhao, Z.~Tong, Y.~He, Y.~Wang, Y.~Wang, and Y.~Qiao, ``{VideoMAE} {V2}: Scaling video masked autoencoders with dual masking,'' in \emph{{IEEE Conf. Comput. Vis. Pattern Recog.}}, 2023, pp. 14\,549--14\,560.

\bibitem{radford2021learning}
A.~Radford, J.~W. Kim, C.~Hallacy, A.~Ramesh, G.~Goh, S.~Agarwal, G.~Sastry, A.~Askell, P.~Mishkin, J.~Clark \emph{et~al.}, ``Learning transferable visual models from natural language supervision,'' in \emph{{Int. Conf. Mach. Learn.}}, 2021, pp. 8748--8763.

\bibitem{wu2024general}
J.~Wu, Y.~Jiang, Q.~Liu, Z.~Yuan, X.~Bai, and S.~Bai, ``General object foundation model for images and videos at scale,'' in \emph{{IEEE Conf. Comput. Vis. Pattern Recog.}}, 2024, pp. 3783--3795.

\bibitem{kirillov2023segment}
A.~Kirillov, E.~Mintun, N.~Ravi, H.~Mao, C.~Rolland, L.~Gustafson, T.~Xiao, S.~Whitehead, A.~C. Berg, W.-Y. Lo \emph{et~al.}, ``Segment anything,'' in \emph{{Int. Conf. Comput. Vis.}}, 2023, pp. 4015--4026.

\bibitem{ravi2024sam}
N.~Ravi, V.~Gabeur, Y.-T. Hu, R.~Hu, C.~Ryali, T.~Ma, H.~Khedr, R.~R{\"a}dle, C.~Rolland, L.~Gustafson \emph{et~al.}, ``{SAM} 2: Segment anything in images and videos,'' \emph{arXiv preprint arXiv:2408.00714}, 2024.

\bibitem{yang2025unimatch}
L.~Yang, Z.~Zhao, and H.~Zhao, ``{UniMatch} {V2}: Pushing the limit of semi-supervised semantic segmentation,'' \emph{{IEEE Trans. Pattern Anal. Mach. Intell.}}, vol.~47, no.~4, pp. 3031--3048, 2025.

\bibitem{benigmim2024collaborating}
Y.~Benigmim, S.~Roy, S.~Essid, V.~Kalogeiton, and S.~Lathuili{\`e}re, ``Collaborating foundation models for domain generalized semantic segmentation,'' in \emph{{IEEE Conf. Comput. Vis. Pattern Recog.}}, 2024, pp. 3108--3119.

\bibitem{xiong2024sam2}
X.~Xiong, Z.~Wu, S.~Tan, W.~Li, F.~Tang, Y.~Chen, S.~Li, J.~Ma, and G.~Li, ``{SAM2}-{UNet}: Segment anything 2 makes strong encoder for natural and medical image segmentation,'' \emph{arXiv preprint arXiv:2408.08870}, 2024.

\bibitem{wu2023medical}
J.~Wu, W.~Ji, Y.~Liu, H.~Fu, M.~Xu, Y.~Xu, and Y.~Jin, ``Medical sam adapter: Adapting segment anything model for medical image segmentation,'' \emph{arXiv preprint arXiv:2304.12620}, 2023.

\bibitem{he2024lightweight}
Y.~He, Z.~Gao, Y.~Li, and Z.~Wang, ``A lightweight multi-modality medical image semantic segmentation network base on the novel unext and wave-mlp,'' \emph{Computerized Medical Imaging and Graphics}, vol. 111, p. 102311, 2024.

\bibitem{chen2014semantic}
L.-C. Chen, G.~Papandreou, I.~Kokkinos, K.~Murphy, and A.~L. Yuille, ``Semantic image segmentation with deep convolutional nets and fully connected crfs,'' \emph{arXiv preprint arXiv:1412.7062}, 2014.

\bibitem{hesham2021pix2pt}
S.~A.~S. Hesham, S.~JinZhou, G.~E.~S. Xin, L.~C. Ping, and L.~Jiang, ``{Pix2Pt} map for transfer-based few-shot learning,'' in \emph{IEEE International Conference on Multimedia \& Expo Workshops}, 2021, pp. 1--6.

\bibitem{chen2017rethinking}
L.-C. Chen, G.~Papandreou, F.~Schroff, and H.~Adam, ``Rethinking atrous convolution for semantic image segmentation,'' \emph{arXiv preprint arXiv:1706.05587}, 2017.

\bibitem{liu2018path}
S.~Liu, L.~Qi, H.~Qin, J.~Shi, and J.~Jia, ``Path aggregation network for instance segmentation,'' in \emph{{IEEE Conf. Comput. Vis. Pattern Recog.}}, 2018, pp. 8759--8768.

\bibitem{bolya2019yolact}
D.~Bolya, C.~Zhou, F.~Xiao, and Y.~J. Lee, ``Yolact: Real-time instance segmentation,'' in \emph{{Int. Conf. Comput. Vis.}}, 2019, pp. 9157--9166.

\bibitem{kirillov2019panoptic}
A.~Kirillov, K.~He, R.~Girshick, C.~Rother, and P.~Doll{\'a}r, ``Panoptic segmentation,'' in \emph{Proceedings of the IEEE/CVF conference on computer vision and pattern recognition}, 2019, pp. 9404--9413.

\bibitem{ainetter2021end}
S.~Ainetter and F.~Fraundorfer, ``End-to-end trainable deep neural network for robotic grasp detection and semantic segmentation from rgb,'' in \emph{IEEE International Conference on Robotics and Automation}, 2021, pp. 13\,452--13\,458.

\bibitem{li2019online}
K.~Li, W.~Tao, and L.~Liu, ``Online semantic object segmentation for vision robot collected video,'' \emph{IEEE Access}, vol.~7, pp. 107\,602--107\,615, 2019.

\bibitem{siam2019video}
M.~Siam, C.~Jiang, S.~Lu, L.~Petrich, M.~Gamal, M.~Elhoseiny, and M.~Jagersand, ``Video object segmentation using teacher-student adaptation in a human robot interaction (hri) setting,'' in \emph{2019 International Conference on Robotics and Automation (ICRA)}.\hskip 1em plus 0.5em minus 0.4em\relax IEEE, 2019, pp. 50--56.

\bibitem{song2020grasping}
S.~Song, A.~Zeng, J.~Lee, and T.~Funkhouser, ``Grasping in the wild: Learning 6dof closed-loop grasping from low-cost demonstrations,'' \emph{IEEE Robotics and Automation Letters}, vol.~5, no.~3, pp. 4978--4985, 2020.

\bibitem{pradeep2022self}
V.~Pradeep, R.~Khemmar, L.~Lecrosnier, Y.~Duchemin, R.~Rossi, and B.~Decoux, ``Self-supervised sidewalk perception using fast video semantic segmentation for robotic wheelchairs in smart mobility,'' \emph{Sensors}, vol.~22, no.~14, p. 5241, 2022.

\bibitem{codevilla2018end}
F.~Codevilla, M.~M{\"u}ller, A.~L{\'o}pez, V.~Koltun, and A.~Dosovitskiy, ``End-to-end driving via conditional imitation learning,'' in \emph{IEEE International Conference on Robotics and Automation}, 2018, pp. 4693--4700.

\bibitem{siam2018comparative}
M.~Siam, M.~Gamal, M.~Abdel-Razek, S.~Yogamani, M.~Jagersand, and H.~Zhang, ``A comparative study of real-time semantic segmentation for autonomous driving,'' in \emph{Proceedings of the IEEE conference on computer vision and pattern recognition workshops}, 2018, pp. 587--597.

\bibitem{muhammad2022vision}
K.~Muhammad, T.~Hussain, H.~Ullah, J.~Del~Ser, M.~Rezaei, N.~Kumar, M.~Hijji, P.~Bellavista, and V.~H.~C. de~Albuquerque, ``Vision-based semantic segmentation in scene understanding for autonomous driving: Recent achievements, challenges, and outlooks,'' \emph{IEEE Transactions on Intelligent Transportation Systems}, vol.~23, no.~12, pp. 22\,694--22\,715, 2022.

\bibitem{xiao2023baseg}
X.~Xiao, Y.~Zhao, F.~Zhang, B.~Luo, L.~Yu, B.~Chen, and C.~Yang, ``Baseg: Boundary aware semantic segmentation for autonomous driving,'' \emph{Neural Networks}, vol. 157, pp. 460--470, 2023.

\bibitem{cciccek20163d}
{\"O}.~{\c{C}}i{\c{c}}ek, A.~Abdulkadir, S.~S. Lienkamp, T.~Brox, and O.~Ronneberger, ``{3D} {U-Net}: Learning dense volumetric segmentation from sparse annotation,'' in \emph{{Med. Image. Comput. Comput. Assist. Interv.}}, 2016, pp. 424--432.

\bibitem{ronneberger2015u}
O.~Ronneberger, P.~Fischer, and T.~Brox, ``{U-Net}: Convolutional networks for biomedical image segmentation,'' in \emph{{Med. Image. Comput. Comput. Assist. Interv.}}, 2015, pp. 234--241.

\bibitem{grammatikopoulou2024spatio}
M.~Grammatikopoulou, R.~Sanchez-Matilla, F.~Bragman, D.~Owen, L.~Culshaw, K.~Kerr, D.~Stoyanov, and I.~Luengo, ``A spatio-temporal network for video semantic segmentation in surgical videos,'' \emph{International Journal of Computer Assisted Radiology and Surgery}, vol.~19, no.~2, pp. 375--382, 2024.

\bibitem{wu2022semi}
H.~Wu, J.~Liu, F.~Xiao, Z.~Wen, L.~Cheng, and J.~Qin, ``Semi-supervised segmentation of echocardiography videos via noise-resilient spatiotemporal semantic calibration and fusion,'' \emph{Medical Image Analysis}, vol.~78, p. 102397, 2022.

\bibitem{li2025semi}
X.~Li, C.~Cui, S.~Shi, H.~Fei, and Y.~Hu, ``Semi-supervised echocardiography video segmentation via adaptive spatio-temporal tensor semantic awareness and memory flow,'' \emph{IEEE Transactions on Medical Imaging}, 2025.

\bibitem{wang2021noisy}
B.~Wang, L.~Li, Y.~Nakashima, R.~Kawasaki, H.~Nagahara, and Y.~Yagi, ``Noisy-lstm: Improving temporal awareness for video semantic segmentation,'' \emph{IEEE Access}, vol.~9, pp. 46\,810--46\,820, 2021.

\bibitem{FCN}
J.~Long, E.~Shelhamer, and T.~Darrell, ``Fully convolutional networks for semantic segmentation,'' in \emph{{IEEE Conf. Comput. Vis. Pattern Recog.}}, 2015, pp. 3431--3440.

\bibitem{mobilenetv2}
M.~Sandler, A.~Howard, M.~Zhu, A.~Zhmoginov, and L.-C. Chen, ``{MobileNetV2}: Inverted residuals and linear bottlenecks,'' in \emph{{IEEE Conf. Comput. Vis. Pattern Recog.}}, 2018, pp. 4510--4520.

\bibitem{DeepLab}
L.-C. Chen, G.~Papandreou, I.~Kokkinos, K.~Murphy, and A.~L. Yuille, ``{DeepLab}: Semantic image segmentation with deep convolutional nets, atrous convolution, and fully connected {CRFs},'' \emph{{IEEE Trans. Pattern Anal. Mach. Intell.}}, vol.~40, no.~4, pp. 834--848, 2017.

\bibitem{PSPNet}
H.~Zhao, J.~Shi, X.~Qi, X.~Wang, and J.~Jia, ``Pyramid scene parsing network,'' in \emph{{IEEE Conf. Comput. Vis. Pattern Recog.}}, 2017, pp. 2881--2890.

\bibitem{yan2020roboseg}
Q.~Yan, S.~Li, C.~Liu, M.~Liu, and Q.~Chen, ``Roboseg: Real-time semantic segmentation on computationally constrained robots,'' \emph{IEEE Transactions on Systems, Man, and Cybernetics: Systems}, vol.~52, no.~3, pp. 1567--1577, 2020.

\bibitem{qiu2017learning}
Z.~Qiu, T.~Yao, and T.~Mei, ``Learning deep spatio-temporal dependence for semantic video segmentation,'' \emph{{IEEE Trans. Multimedia}}, vol.~20, no.~4, pp. 939--949, 2017.

\bibitem{MRCFA}
G.~Sun, Y.~Liu, H.~Tang, A.~Chhatkuli, L.~Zhang, and L.~Van~Gool, ``Mining relations among cross-frame affinities for video semantic segmentation,'' in \emph{{Eur. Conf. Comput. Vis.}}, 2022, pp. 522--539.

\bibitem{li2018low}
Y.~Li, J.~Shi, and D.~Lin, ``Low-latency video semantic segmentation,'' in \emph{{IEEE Conf. Comput. Vis. Pattern Recog.}}, 2018, pp. 5997--6005.

\bibitem{paul2020efficient}
M.~Paul, C.~Mayer, L.~V. Gool, and R.~Timofte, ``Efficient video semantic segmentation with labels propagation and refinement,'' in \emph{Proceedings of the IEEE/CVF Winter Conference on Applications of Computer Vision}, 2020, pp. 2873--2882.

\bibitem{hu2020temporally}
P.~Hu, F.~Caba, O.~Wang, Z.~Lin, S.~Sclaroff, and F.~Perazzi, ``Temporally distributed networks for fast video semantic segmentation,'' in \emph{{IEEE Conf. Comput. Vis. Pattern Recog.}}, 2020, pp. 8818--8827.

\bibitem{shelhamer2016clockwork}
E.~Shelhamer, K.~Rakelly, J.~Hoffman, and T.~Darrell, ``Clockwork convnets for video semantic segmentation,'' in \emph{{Eur. Conf. Comput. Vis.}}\hskip 1em plus 0.5em minus 0.4em\relax Springer, 2016, pp. 852--868.

\bibitem{zhu2019improving}
Y.~Zhu, K.~Sapra, F.~A. Reda, K.~J. Shih, S.~Newsam, A.~Tao, and B.~Catanzaro, ``Improving semantic segmentation via video propagation and label relaxation,'' in \emph{{IEEE Conf. Comput. Vis. Pattern Recog.}}, 2019, pp. 8856--8865.

\bibitem{zhou2022survey}
T.~Zhou, F.~Porikli, D.~J. Crandall, L.~Van~Gool, and W.~Wang, ``A survey on deep learning technique for video segmentation,'' \emph{{IEEE Trans. Pattern Anal. Mach. Intell.}}, vol.~45, no.~6, pp. 7099--7122, 2022.

\bibitem{gao2023deep}
M.~Gao, F.~Zheng, J.~J. Yu, C.~Shan, G.~Ding, and J.~Han, ``Deep learning for video object segmentation: a review,'' \emph{Artificial Intelligence Review}, vol.~56, no.~1, pp. 457--531, 2023.

\bibitem{pan2024moda}
F.~Pan, X.~Yin, S.~Lee, A.~Niu, S.~Yoon, and I.~S. Kweon, ``Moda: Leveraging motion priors from videos for advancing unsupervised domain adaptation in semantic segmentation,'' in \emph{{IEEE Conf. Comput. Vis. Pattern Recog.}}, 2024, pp. 2649--2658.

\bibitem{guo2024vanishing}
D.~Guo, D.-P. Fan, T.~Lu, C.~Sakaridis, and L.~Van~Gool, ``Vanishing-point-guided video semantic segmentation of driving scenes,'' in \emph{CVPR}, 2024, pp. 3544--3553.

\bibitem{mai2024pay}
H.~Mai, R.~Sun, Y.~Wang, T.~Zhang, and F.~Wu, ``Pay attention to target: Relation-aware temporal consistency for domain adaptive video semantic segmentation,'' in \emph{{AAAI Conf. Artif. Intell.}}, vol.~38, no.~5, 2024, pp. 4162--4170.

\bibitem{MPVSS}
Y.~Weng, M.~Han, H.~He, M.~Li, L.~Yao, X.~Chang, and B.~Zhuang, ``Mask propagation for efficient video semantic segmentation,'' in \emph{{Adv. Neural Inform. Process. Syst.}}, 2024, pp. 7170--7183.

\bibitem{gao2023exploit}
Y.~Gao, Z.~Wang, J.~Zhuang, Y.~Zhang, and J.~Li, ``Exploit domain-robust optical flow in domain adaptive video semantic segmentation,'' in \emph{{AAAI Conf. Artif. Intell.}}, vol.~37, no.~1, 2023, pp. 641--649.

\bibitem{TV3S}
S.~A.~S. Hesham, Y.~Liu, G.~Sun, H.~Ding, J.~Yang, E.~Konukoglu, X.~Geng, and X.~Jiang, ``Exploiting temporal state space sharing for video semantic segmentation,'' in \emph{{IEEE Conf. Comput. Vis. Pattern Recog.}}, 2025, pp. 24\,211--24\,221.

\bibitem{lee2025cavis}
S.~Lee, J.~Seo, K.~Han, M.~Choi, and S.~Im, ``Cavis: Context-aware video instance segmentation,'' in \emph{{IEEE Conf. Comput. Vis. Pattern Recog.}}, 2025, pp. 4507--4517.

\bibitem{zhang2023dvis}
T.~Zhang, X.~Tian, Y.~Wu, S.~Ji, X.~Wang, Y.~Zhang, and P.~Wan, ``Dvis: Decoupled video instance segmentation framework,'' in \emph{Proceedings of the IEEE/CVF International Conference on Computer Vision}, 2023, pp. 1282--1291.

\bibitem{zhang2025dvis++}
T.~Zhang, X.~Tian, Y.~Zhou, S.~Ji, X.~Wang, X.~Tao, Y.~Zhang, P.~Wan, Z.~Wang, and Y.~Wu, ``Dvis++: Improved decoupled framework for universal video segmentation,'' \emph{IEEE Transactions on Pattern Analysis and Machine Intelligence}, 2025.

\bibitem{bommasani2021opportunities}
R.~Bommasani, D.~A. Hudson, E.~Adeli, R.~Altman, S.~Arora, S.~von Arx, M.~S. Bernstein, J.~Bohg, A.~Bosselut, E.~Brunskill \emph{et~al.}, ``On the opportunities and risks of foundation models,'' \emph{arXiv preprint arXiv:2108.07258}, 2021.

\bibitem{devlin2019bert}
J.~Devlin, M.-W. Chang, K.~Lee, and K.~Toutanova, ``{BERT}: Pre-training of deep bidirectional transformers for language understanding,'' in \emph{Proceedings of the 2019 Conference of the North American Chapter of the Association for Computational Linguistics: Human Language Technologies}, 2019, pp. 4171--4186.

\bibitem{achiam2023gpt}
J.~Achiam, S.~Adler, S.~Agarwal, L.~Ahmad, I.~Akkaya, F.~L. Aleman, D.~Almeida, J.~Altenschmidt, S.~Altman, S.~Anadkat \emph{et~al.}, ``{GPT}-4 technical report,'' \emph{arXiv preprint arXiv:2303.08774}, 2023.

\bibitem{hurst2024gpt}
A.~Hurst, A.~Lerer, A.~P. Goucher, A.~Perelman, A.~Ramesh, A.~Clark, A.~Ostrow, A.~Welihinda, A.~Hayes, A.~Radford \emph{et~al.}, ``{GPT}-4o system card,'' \emph{arXiv preprint arXiv:2410.21276}, 2024.

\bibitem{jaech2024openai}
A.~Jaech, A.~Kalai, A.~Lerer, A.~Richardson, A.~El-Kishky, A.~Low, A.~Helyar, A.~Madry, A.~Beutel, A.~Carney \emph{et~al.}, ``{OpenAI} o1 system card,'' \emph{arXiv preprint arXiv:2412.16720}, 2024.

\bibitem{jia2021scaling}
C.~Jia, Y.~Yang, Y.~Xia, Y.-T. Chen, Z.~Parekh, H.~Pham, Q.~Le, Y.-H. Sung, Z.~Li, and T.~Duerig, ``Scaling up visual and vision-language representation learning with noisy text supervision,'' in \emph{{Int. Conf. Mach. Learn.}}, 2021, pp. 4904--4916.

\bibitem{yang2023track}
J.~Yang, M.~Gao, Z.~Li, S.~Gao, F.~Wang, and F.~Zheng, ``Track anything: Segment anything meets videos,'' \emph{arXiv preprint arXiv:2304.11968}, 2023.

\bibitem{cheng2023segment}
Y.~Cheng, L.~Li, Y.~Xu, X.~Li, Z.~Yang, W.~Wang, and Y.~Yang, ``Segment and track anything,'' \emph{arXiv preprint arXiv:2305.06558}, 2023.

\bibitem{he2023scalable}
H.~He, J.~Zhang, M.~Xu, J.~Liu, B.~Du, and D.~Tao, ``Scalable mask annotation for video text spotting,'' \emph{arXiv preprint arXiv:2305.01443}, 2023.

\bibitem{lu2023can}
Z.~Lu, Z.~Xiao, J.~Bai, Z.~Xiong, and X.~Wang, ``Can {SAM} boost video super-resolution?'' \emph{arXiv preprint arXiv:2305.06524}, 2023.

\bibitem{wang2024disco}
T.~Wang, L.~Li, K.~Lin, Y.~Zhai, C.-C. Lin, Z.~Yang, H.~Zhang, Z.~Liu, and L.~Wang, ``{DisCo}: Disentangled control for realistic human dance generation,'' in \emph{{IEEE Conf. Comput. Vis. Pattern Recog.}}, 2024, pp. 9326--9336.

\bibitem{qin2023dancing}
B.~Qin, W.~Ye, Q.~Yu, S.~Tang, and Y.~Zhuang, ``Dancing {Avatar}: Pose and text-guided human motion videos synthesis with image diffusion model,'' \emph{arXiv preprint arXiv:2308.07749}, 2023.

\bibitem{yang2023sam3d}
Y.~Yang, X.~Wu, T.~He, H.~Zhao, and X.~Liu, ``{SAM3D}: Segment anything in {3D} scenes,'' \emph{arXiv preprint arXiv:2306.03908}, 2023.

\bibitem{dong2023leveraging}
S.~Dong, F.~Liu, and G.~Lin, ``Leveraging large-scale pretrained vision foundation models for label-efficient {3D} point cloud segmentation,'' \emph{arXiv preprint arXiv:2311.01989}, 2023.

\bibitem{lai2023detect}
Y.~Lai, Z.~Luo, and Z.~Yu, ``Detect any deepfakes: Segment anything meets face forgery detection and localization,'' in \emph{Chinese Conference on Biometric Recognition}, 2023, pp. 180--190.

\bibitem{wu2023cvpr}
J.~Z. Wu, X.~Li, D.~Gao, Z.~Dong, J.~Bai, A.~Singh, X.~Xiang, Y.~Li, Z.~Huang, Y.~Sun \emph{et~al.}, ``{CVPR} 2023 text guided video editing competition,'' \emph{arXiv preprint arXiv:2310.16003}, 2023.

\bibitem{yin2023or}
Y.~Yin, Z.~Fu, F.~Yang, and G.~Lin, ``{OR}-{NeRF}: Object removing from {3D} scenes guided by multiview segmentation with neural radiance fields,'' \emph{arXiv preprint arXiv:2305.10503}, 2023.

\bibitem{zhao2023fast}
X.~Zhao, W.~Ding, Y.~An, Y.~Du, T.~Yu, M.~Li, M.~Tang, and J.~Wang, ``Fast segment anything,'' \emph{arXiv preprint arXiv:2306.12156}, 2023.

\bibitem{chu2024zero}
W.-H. Chu, A.~W. Harley, P.~Tokmakov, A.~Dave, L.~Guibas, and K.~Fragkiadaki, ``Zero-shot open-vocabulary tracking with large pre-trained models,'' in \emph{IEEE International Conference on Robotics and Automation}, 2024, pp. 4916--4923.

\bibitem{mo2023av}
S.~Mo and Y.~Tian, ``{AV}-{SAM}: Segment anything model meets audio-visual localization and segmentation,'' \emph{arXiv preprint arXiv:2305.01836}, 2023.

\bibitem{yue2024surgicalsam}
W.~Yue, J.~Zhang, K.~Hu, Y.~Xia, J.~Luo, and Z.~Wang, ``Surgical{SAM}: Efficient class promptable surgical instrument segmentation,'' in \emph{{AAAI Conf. Artif. Intell.}}, vol.~38, no.~7, 2024, pp. 6890--6898.

\bibitem{benjdira2023rosgpt_vision}
B.~Benjdira, A.~Koubaa, and A.~M. Ali, ``{ROSGPT}\_{Vision}: Commanding robots using only language models' prompts,'' \emph{arXiv preprint arXiv:2308.11236}, 2023.

\bibitem{fu2024sam}
C.~Fu, L.~Yao, H.~Zuo, G.~Zheng, and J.~Pan, ``{SAM}-{DA}: {UAV} tracks anything at night with {SAM}-powered domain adaptation,'' in \emph{International Conference on Advanced Robotics and Mechatronics}, 2024, pp. 31--38.

\bibitem{xie2021Segformer}
E.~Xie, W.~Wang, Z.~Yu, A.~Anandkumar, J.~M. Alvarez, and P.~Luo, ``{SegFormer}: Simple and efficient design for semantic segmentation with transformers,'' in \emph{{Adv. Neural Inform. Process. Syst.}}, 2021, pp. 12\,077--12\,090.

\bibitem{miao2022large}
J.~Miao, X.~Wang, Y.~Wu, W.~Li, X.~Zhang, Y.~Wei, and Y.~Yang, ``Large-scale video panoptic segmentation in the wild: A benchmark,'' in \emph{{IEEE Conf. Comput. Vis. Pattern Recog.}}, 2022, pp. 21\,033--21\,043.

\bibitem{chen2024sam2}
T.~Chen, A.~Lu, L.~Zhu, C.~Ding, C.~Yu, D.~Ji, Z.~Li, L.~Sun, P.~Mao, and Y.~Zang, ``{SAM2}-{Adapter}: Evaluating \& adapting segment anything 2 in downstream tasks: Camouflage, shadow, medical image segmentation, and more,'' \emph{arXiv preprint arXiv:2408.04579}, 2024.

\bibitem{chen2025samcp}
\BIBentryALTinterwordspacing
P.~Chen, L.~Xie, X.~Huo, X.~Yu, X.~ZHANG, Y.~Sun, Z.~Han, and Q.~Tian, ``{SAM}-{CP}: Marrying {SAM} with composable prompts for versatile segmentation,'' in \emph{{Int. Conf. Learn. Represent.}}, 2025. [Online]. Available: \url{https://openreview.net/forum?id=UiEjzBRYeI}
\BIBentrySTDinterwordspacing

\end{thebibliography}

\bigskip

\newcommand{\AddPhoto}[1]{\includegraphics[width=.86in,clip,keepaspectratio]{#1}}

\begin{biography}{\AddPhoto{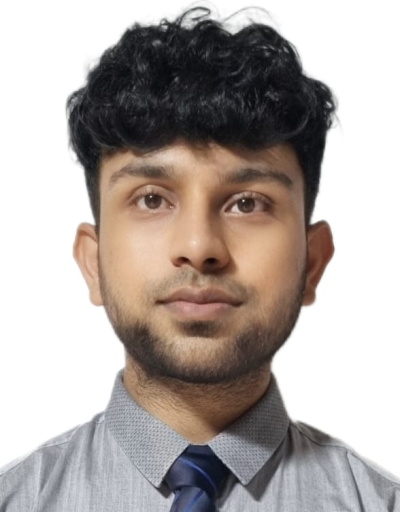}}{\textbf{Syed Hesham} is currently a Ph.D. candidate in Electrical and Electronic Engineering at Nanyang Technological University (NTU) in Singapore, and is also an A*STAR Computing and Information Science (ACIS) Research Scholar working under the Institute for Infocomm Research (I2R). He earned his Bachelor's degree in Electrical and Electronic Engineering from NTU in 2023.

His research is at the intersection of deep learning and computer vision, with a particular focus on advancing effective and label-efficient visual perception models. 

E-mail: syedhesh002@e.ntu.edu.sg

ORCID iD: 0009-0000-9589-3056}
\end{biography}

\begin{biography}{\AddPhoto{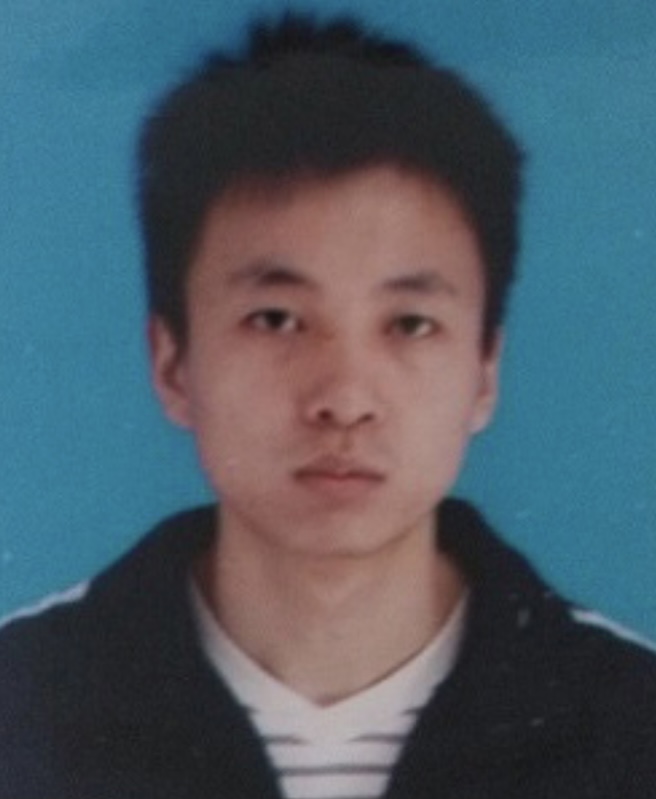}}{\textbf{Yun Liu} received his B.E. and Ph.D. degrees from Nankai University in 2016 and 2020, respectively. Then, he worked with Prof. Luc Van Gool as a postdoctoral scholar at the Computer Vision Lab, ETH Zurich, Switzerland. After that, he worked as a senior scientist at the Institute for Infocomm Research (I2R), A*STAR, Singapore. Now, he is a professor at the College of Computer Science, Nankai University.

His research interests include computer vision and machine learning (especially deep learning).

E-mail: liuyun@nankai.edu.cn

ORCID iD: 0000-0001-6143-0264}
\end{biography}

\begin{biography}{\AddPhoto{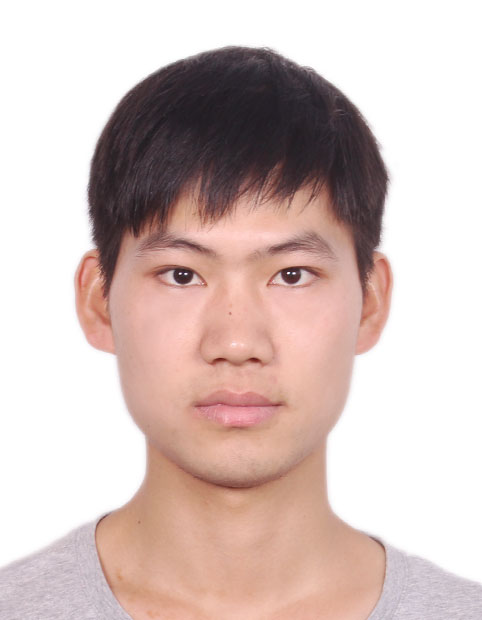}}{\textbf{Guolei Sun} 
received his Ph.D. degree at ETH Zurich, Switzerland, in Prof. Luc Van Gool's Computer Vision Lab in Jan 2024.  Before that, he got his master's degree in computer science from the King Abdullah University of Science and Technology (KAUST), in 2018. He is currently a postdoctoral researcher at the Computer Vision Lab, ETH Zurich. From 2018 to 2019, he worked as a research engineer with the Inception Institute of Artificial Intelligence, UAE.

His research interests include deep learning for video understanding, semantic/instance segmentation, object counting, and weakly supervised learning.

E-mail: guolei.sun@vision.ee.ethz.ch

ORCID iD: 0000-0001-8667-9656}
\end{biography}

\begin{biography}{\AddPhoto{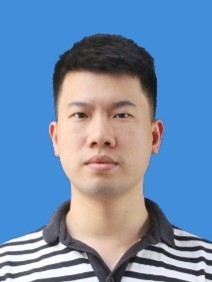}}{\textbf{Jing Yang}
received his Ph.D. degree in mechanical and electronic engineering from Guizhou University in 2020. From August 2018 to September 2019, he was awarded a scholarship by the China Scholarship Council (CSC) under the State Scholarship Fund to pursue his study with Oklahoma State University, as a Joint Ph.D. Student with the School of Computer Science and Technology, where he joined the Guoliang Fan's Group. From October 2022 to October 2023, he was a visiting scholar studying in the team of Professor Guo Minyi (IEEE Fellow) from Shanghai Jiao Tong University. He is currently an assistant professor with the State Key Laboratory of Public Big Data, Guizhou University, China.

His research interests include high-performance computing, task scheduling in various architectures, and open-domain visual learning.

E-mail: jyang23@gzu.edu.cn

ORCID iD: 0000-0003-1915-9487}
\end{biography}

\begin{biography}{\AddPhoto{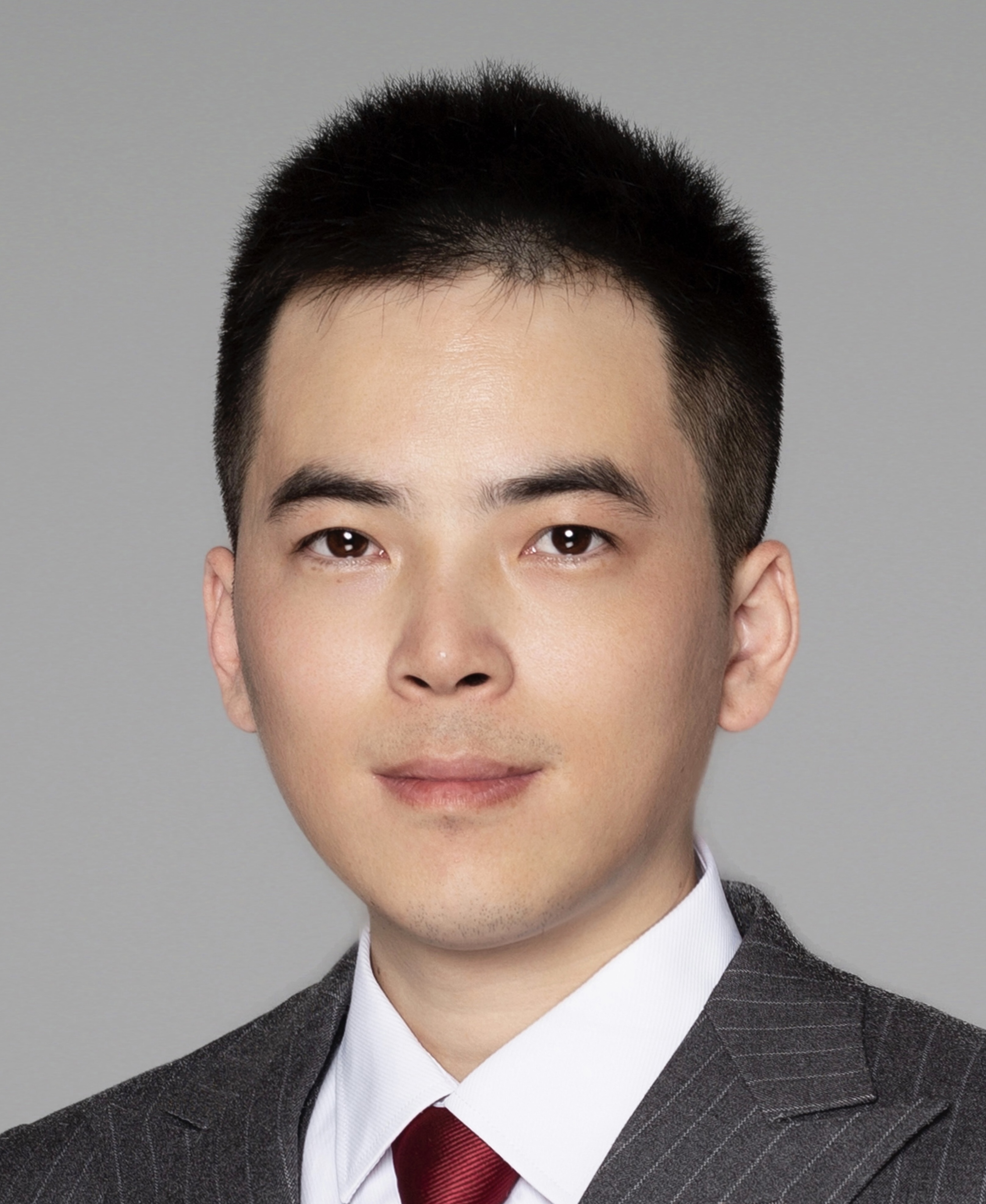}}{\textbf{Henghui
Ding} 
received the B.E. degree from Xi'an Jiaotong University, Xi'an, China, in 2016. He received his Ph.D. degree from Nanyang Technological University (NTU), Singapore, in 2020. He was a Research Scientist at ByteDance, Beijing, China, and a Post-Doctoral Researcher at ETH and NTU. He is currently a tenure-track professor at Fudan University, Shanghai, China. He serves as an Associate Editor for IET Computer Vision and Visual Intelligence and serves/has served as the Area Chair of CVPR'24, NeurIPS'24, ICLR'25, ACM MM'24, and BMVC'24 and an SPC Member of AAAI'(22-25) and IJCAI'(23-24).

His research interests include computer vision and machine learning.

E-mail: hhding@fudan.edu.cn

ORCID iD: 0000-0003-4868-6526}
\end{biography}

\begin{biography}{\AddPhoto{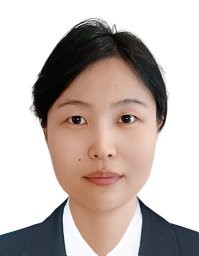}}{\textbf{Xue Geng} 
received her B.E. degree in computer science from Northeastern University, China, in 2012, and her Ph.D. degree in computer science from National University of Singapore (NUS), Singapore, in 2017. Currently, she is a senior scientist with the Institute for Infocomm Research (I2R), A*STAR, Singapore.

Her research interests include model compression and efficient machine learning.

E-mail: snownus@gmail.com

ORCID iD: 0000-0002-2594-9648}
\end{biography}

\begin{biography}{\AddPhoto{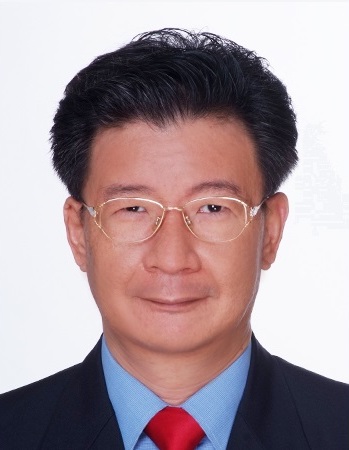}}{\textbf{Xudong Jiang} 
(Fellow, IEEE) received the B.E. and M.E. degrees from the University of Electronic Science and Technology of China (UESTC), Chengdu, China, in 1983 and 1986, respectively, and the Ph.D. degree from Helmut Schmidt University, Hamburg, Germany, in 1997. From 1998 to 2004, he was with the Institute for Infocomm Research, A*STAR, Singapore, as a Lead Scientist, and the Head of the Biometrics Laboratory. He joined Nanyang Technological University (NTU), Singapore, as a Faculty Member in 2004, where he was the Director of the Center for Information Security from 2005 to 2011. He is currently a professor with the School of EEE, NTU, and the Director of the Centre for Information Sciences and Systems, Singapore. He has authored more than 200 articles, with over 60 articles in IEEE journals and 30 papers in top conferences such as CVPR/ICCV/ECCV/AAAI/ICLR. He has served as an Associate Editor for IEEE Signal Processing Letters (IEEE SPL) and IEEE Transactions on Image Processing (IEEE T-IP). Currently, he serves as the Senior Area Editor for IEEE Transactions on Image Processing and the Editor-in-Chief of IET Biometrics.

His research interests include computer vision, machine learning, pattern recognition, image processing, and biometrics.

E-mail: exdjiang@ntu.edu.sg

ORCID iD: 0000-0002-9104-2315}
\end{biography}

\end{document}